\documentclass{article}
\usepackage{smc2019}
\usepackage{times}
\usepackage{ifpdf}
\usepackage[english]{babel}
\usepackage{cite}
\usepackage{adjustbox, float, longtable, wrapfig,url}
\usepackage{fancyhdr,comment}
\usepackage{amsmath, soul}
\usepackage{amssymb}
\usepackage[framed,amsmath,thmmarks]{ntheorem}

\def\papertitle{End-To-End Dilated Variational Autoencoder with Bottleneck Discriminative Loss for Sound Morphing - A Preliminary Study}
\def\firstauthor{Matteo Lionello}
\def\secondauthor{Hendrik Purwins}

\newif\ifpdf
\ifx\pdfoutput\relax
\else
   \ifcase\pdfoutput
      \pdffalse
   \else
      \pdftrue
\fi

\ifpdf 
  \usepackage[pdftex,
    pdftitle={\papertitle},
    pdfauthor={\firstauthor, \secondauthor},
    bookmarksnumbered, 
    pdfstartview=XYZ 
   ]{hyperref}

  \usepackage[figure,table]{hypcap}

\else 
  \usepackage[dvips,
    bookmarksnumbered, 
    pdfstartview=XYZ 
  ]{hyperref}  

  \usepackage[dvips]{epsfig,graphicx}
  \DeclareGraphicsExtensions{.eps}

  \usepackage[figure,table]{hypcap}
\fi

\hypersetup{
    colorlinks,%
    citecolor=black,%
    filecolor=black,%
    linkcolor=black,%
    urlcolor=black
}

\title{\papertitle}

\author{\setlength{\tabcolsep}{32pt} \begin{tabular}{c c}
    Matteo Lionello & Hendrik Purwins \\
    University College London & Accenture \\ {\tt \href{mailto:lionello.matteo@gmail.com}{matteo.lionello.18@ucl.ac.uk}} &  {\tt \href{mailto:hendrik.purwins@accenture.com}{hendrik.purwins@accenture.com}}
     \end{tabular}}

\def\thebibliography#1{
  \section{References}\list
  {[\arabic{enumi}]}{
  \settowidth\labelwidth{[#1]}\leftmargin 1em
  \advance\leftmargin\labelsep
  \usecounter{enumi}
  }
  \def\newblock{\hskip .01em plus .01em minus .01em}
  \sloppy\clubpenalty4000\widowpenalty4000
  \sfcode`\.=1000\relax
}

\begin{document}
\capstartfalse
\maketitle
\begin{abstract} 
We present a preliminary study\footnote{Both data and experiments reported in this paper were part of Matteo Lionello's master's thesis conducted in Spring 2018 at Aalborg University Copenhagen\cite{thesis}} on an end-to-end variational autoencoder (VAE) for sound morphing. Two VAE variants are compared: VAE with dilation layers (DC-VAE) and VAE only with regular convolutional layers (CC-VAE).
We combine the following loss functions: 1) the  time-domain mean-squared error for reconstructing the input signal, 2) the Kullback-Leibler divergence to the standard normal distribution in the bottleneck layer, and 3) the classification loss calculated from the bottleneck representation.

On a database of spoken digits, we use 1-nearest neighbor classification to show that the sound classes separate in the bottleneck layer.  We introduce the Mel-frequency cepstrum coefficient dynamic time warping (MFCC-DTW) deviation as a measure of how well the VAE decoder projects the class center in the latent (bottleneck) layer to the center of the sounds of that class in the audio domain. In terms of MFCC-DTW deviation and 1-NN classification,  DC-VAE outperforms CC-VAE. These results for our parametrization and our dataset indicate that DC-VAE is more suitable for sound morphing than CC-VAE, since the DC-VAE decoder better preserves the topology when mapping from the audio domain to the latent space. Examples are given both for morphing spoken digits and drum sounds.
\end{abstract}

\section{Introduction}
Audio synthesis comprises a set of methods to generate sounds usually from scratch. It finds its main applications in text-to-speech generation - vocoders - and music production - synthesizers.  Speech synthesis is usually done by formant synthesis, concatenative synthesis or statistical parametric synthesis. There exist multiple strategies to create sounds to emulate existing music instruments or to create new sounds of instruments that actually do not exist in the real world. The main techniques consist of granular sythesis\cite{Xenakis}, frequency modulation\cite{chowning}, subtractive synthesis, additive synthesis, sampled synthesis, wavetables\cite{wavetable}, modelling physical systems\cite{x}.
Sound morphing is an audio transformation technique which generates new sounds by gradually mixing  different properties of the existing ones. Sound morphing can be performed by spectral modelling synthesis \cite{serra} which models time varying spectra by means of sinusoidal decomposition and noise components.
Research in artificial multi-layers neural networks has provided new tools to generate sound for both speech and sound synthesis\cite{purwins2019deep}.

Convolution neural networks (CNNs) have been used to synthesize and manipulate music and speech, in conjunction with generative adversarial networks \cite{donahue_synthesizing_2018}, stacked dilated convolutions \cite{nsynth, wavenet} or stacked recurrent layers \cite{mehri_samplernn2016}.
When using an autoencoder in which different sound classes  map to different clusters  in  the latent (bottleneck) layer, interpolations between those clusters can yield morphed sounds. NSynth\cite{nsynth} generates sound using the WaveNet \cite{wavenet} as a decoder, conditioned on the latent representation of an autoencoder. Compared to NSynth, our approach is less computationally heavy. In addition, we use a  variational autoencoder (VAE, \cite{Variational0}). In \cite{hsu2017learning}, the VAE is used to morph speech snippets represented as spectrograms. 
Opposed to a spectrogram representation, we use raw audio as input for our VAE which spares us the difficulty to generate phases from the magnitude spectrum.

The goal of this work is to create a model for generating new sounds that can be easily controlled by changing the intermediate representation of a model. The model is built to learn to reconstruct a set of audio files by extracting a low-dimensional representation from them. In the current work, this is achieved by means of Deep Learning methods based on autoencoders which learn how to represent a given dataset by extracting the features necessary to reconstruct it. The feature extraction is done by the autoencoder by projecting the audio into a low-dimensional space defined as a probability space. Once the training process of the model ends, it makes it possible to explore the latent space and to create a new representation from it between two regions where two different kinds of sounds cluster. By reconstructing the set of points sampled from a path starting from one region and ending in the other one, a morphing between the two sounds will be achieved.

This study shows an application of variational autoencoder to the audio domain. Audio transformation by means of autoencoders has already been used to upsample an audio signal\cite{gansSRNN}. 
However such a basic approach is not very flexible to be used e.g. for sound morphing with changing frequencies. By applying a variational inference, we provide a organized latent representation that we can control to generate the output of the model.

Some pilot experiments we conducted, showed that autoencoders based on dilation and convolution layers store at their bottleneck a lower sampled signal representation of the input which maintains the same frequency of the input. To avoid aliasing effects due to the compression rate - indeed if the signal stored at the bottleneck is a low sample representation of the original input the frequencies which can be represented at the bottleneck are lower too - the autoencoders learn to modulate the amplitude of the latent signal in the bottleneck to reconstruct those whose frequency is larger than the frequency limit given by  Nyquist's theorem. 

To do sound morphing, we need to free the latent space from its temporal sequence representation by forcing the latent representation to respect additional requirements.
We propose a variant of the variational autoencoder \cite{Variational0, Variational1} yielding a model which provides an easily tractable latent representation, from which we can generate and morph sounds in a meaningful way.

\section{Related Work}
In the last couple of years, a rapid evolution of methods to generate data by means of Deep Learning techniques has been observed, providing interesting perspectives in how to represent and generate audio and images. As seen by the large amount of projects which are based on it, the most relevant work achieved in audio domain by means of Deep Learning tools is the introduction of WaveNet model \cite{wavenet}.

\label{wavenet}
\textbf{WaveNet}\cite{wavenet} is a deep learning model that uses dilation layers to extrapolate inter time dependencies within one sound. The model can be used to generate music and speech controlled by a conditioning vector.
Similarly to two previous projects, PixelRNN\cite{pixelcnn} and PixelCNN\cite{pixelrnn}, WaveNet sequentially predicts  each output sample into 128 channels.
By doing so, each sample is linked to a 128 bins distribution of possible values calculated by the neural network.
The neural network compares the predicted probability distribution of the next sample to the one-hot vector of the original following sample. 
The price for its good performance is the computational expense, both in time and resources, which makes WaveNet in this form not usable for real-time tasks.
To increase the length of the receptive field by maintaining filter of short length and not increasing excessively the amount of layers, a dilation factor has been introduced to let the filter between two layers skip consecutive samples. This strategy allows the receptive field to grow exponentially with respect to the depth of the network.
The layers used for our current work have been modelled similarly to the dilation layers used in WaveNet. Stack of parallel causal layers were introduced to extrapolate features that are spread distributed along the signal.

With \textbf{Parallel WaveNet}\cite{parallelwavenet}, published in 2017, the algorithm has been accelerated to be used in real-time applications. Parallel WaveNet introduces the combination of three different losses. The first one is the Probability Density Distillation which is calculated as the Kullback-Leibler divergence between the output of a "Teacher" WaveNet network already trained and a "Student" WaveNet network. The two networks are linked in cascade where the "Student" network plays the role of the generator and the "Teacher" plays the role of the discriminator. The other two losses were introduced to improve the quality of the output: a power loss to ensure that the output
of the model has frequency bands similar to the ones of the target sound, and a perceptual loss to classify the phonemes correctly.
The shorter time used by Parallel WaveNet for the predictions is a partial advantage since the "Teacher" WaveNet network has to be fully trained sample-by-sample previously.

\textbf{NSynth}\label{rel:nsynth}\cite{nsynth} is a neural synthesizer that consists of an autoencoder whose components are similar to what is used in WaveNet. They showed that WaveNet needs persistent external conditioning to generate stable outputs. NSynth aimed at introducing a model which overcomes this problem. 
To avoid the external conditioning for each set of 512 samples (corresponding to 32 ms of the original audio), the model decodes a chunk into an embedding space of 16 dimensions ($\times 32$ compression) whose average is used as conditioning for a WaveNet decoder network whose input is the same original audio.
Some differences between our approach and NSynth have to be remarked: we propose an autoencoder that uses variational inference and a classification loss in the bottleneck to represent one entire sound as one 20-dimensional point. NSynth uses a 16 dimensional vector for every 32ms of each audio. This choice allows NSynth to have a temporal representation in the bottleneck. This achievement can be easily reached by using raw autoencoders, while the goal of our work was to build a latent representation not depending on a temporal representation. A second difference is that the NSynth uses a WaveNet decoder which regenerates the output sample by sample by using the quantised distribution for each sample to be predicted, while the final layer of the model hereby proposed is composed of a convolutional layer allowing a simpler training. 
Moreover a weak point of NSynth is the inherited training sample-by-sample  of the WaveNet, making the training process computationally heavy. One main goal of the current project is to develop a simple structure that is more easily tractable, freeing the autoencoder from the quantization process.

A current competitor of the framework of WaveNet is \textbf{SING}\cite{sing} that introduces a model based on a Long-Short Term Memory\cite{lstm} sequence generator which last hidden state is decoded by a convolutional network.

\subsection{Probabilistic generative models}
Probabilistic generative modelling is used in machine learning to generate data whose probability representation is learned to be similar to the dataset used in the training.
On the contrary, the goal in discriminative models\cite{bishop_disc} which learn classification tasks by predicting posterior probabilities of class assignments after learning the likelihood features vs classes from the training dataset. On the other hand, generative models try to learn the join probability distribution\cite{generative, NIPS2001_2020} that describes the dataset and then to generate new data by sampling the distribution obtained.
Variational Autoencoders (VAE) and Generative Adversarials Networks (GANs) are one of the most common method in machine learning to achieve this goal. 
This is the case, for instance, of a music style transfer with an end-to-end approach by means of GANs structured through WaveNet-based encoders and decoders for conditioning the network\cite{fb}. By comparing directly GANs and VAE, the latent space used in GANs is sampled from white noise and used directly as input to the networks. This strategy makes the mapping of the latent space  much more difficult, involving a second network to learn separately the inverse function. Many other ways to introduce an inverse mapping are currently under investigation\cite{invgans,semgans,bigans}. VAEs, on the other side, map directly the input into the latent space and then back to the original feature space. This strategy allows to adapt the hidden space according to arbitrary criteria and to shape it in more human-interpretable space.

Autoencoders are multi-layer neural networks whose goal is to reconstruct the input after a dimensionality reduction computed within the network. The most basic autoencoder (excluding advanced variations of it such as sparse autoencoders \cite{sparse}) is made of two components: the encoder and the decoder. The encoder consists of a sequence of layers which sequentially decrease the dimension of their outputs. Opposite to this, the decoder is a stack of layers whose dimensions increase until they reach the same size of input dimension. The bottleneck of this architecture corresponds to the layer at the end of the encoder and its dimension is the smallest among the layers of the network. For optimal reconstruction after the bottleneck layer, the autoencoder enforces an efficient latent representation of the input in the bottleneck layer.

An autoencoder can be mathematically represented as a concatenation of encoder $\theta$ and decoder $\phi$.
$\theta$ maps from the input domain $\mathcal{X}\subset \mathbb{R}^{q}$ (audio in our case) to the latent space (bottleneck) $\mathcal{Z}\subset \mathbb{R}^{s}$ (with smaller dimension $s$ than the input space):
$\theta: \mathcal{X} \rightarrow \mathcal{Z}$.
The decoder $\phi$ maps back from the latent space $\mathcal{Z}$ to a subspace of the input domain $\mathcal{Y} \subset \mathbb{R}^{q}:$  $\phi: \mathcal{Z} \rightarrow \mathcal{Y}$. 
During training, the autoencoder aims at achieving $\theta,\phi = argmin_{\theta,\phi}\{L(X,(\theta\circ \phi )(X))\}$ where $L$ is a loss function $\mathcal{X}\times\mathcal{Y}\rightarrow \mathbb{R}$ that calculate the dissimilarities between its arguments. $L$ can be the reconstruction error for example.
For an original sample $\vec{x}_{i} \in \mathcal{X}$, $(\theta\circ \phi) (\vec{x}_{i}) = \vec{y}_{i} \in \mathcal{Y}$ is the reconstruction of the datapoint
$\vec{x}_{i}$, while $\vec{z}_{i} = \theta (\vec{x}_{i}) \in \mathcal{Z}$ is the latent - or coded - representation of $\vec{x}_{i}$. 
 $\phi(\vec{x}_{i}\in \mathcal{X})$ is the latent representation of the data $\vec{x}_{i}$ computed by the encoder.

The \textbf{Variational Autoencoder}\cite{Variational0,Variational1,tut_var,tut_var1} is a method which uses the latent representation of the autoencoder to approximate a probability distribution. This encourages the samples to be represented in a meaningful way in the latent space distributed according to some prior. Some studies \cite{vaeq} show that it is also possible to infer the latent space with discrete representation by nearest neighbour calculation before fitting the sampled random variable representing the sample into the decoder. The objective function of a Variational Autoencoder is the sum of the generation loss (the expected likelihood calculated in our case through a L2 loss) $\mathcal{L}_{Likelihood}$ and a regularizing term (the Kullback-Leibler divergence) $\mathcal{L}_{KL}$:
\begin{equation}\label{vaeobj}
\|\phi(\theta(x))-x\|_{2}^{2}+KL(\mathcal{N}(\theta_{\mu}(x),\theta_{\Sigma}(x))\|\mathcal{N}(0,I)) \end{equation}
The last layer of the encoder consists of two parallel dense layers: $\theta_{\mu}$ and $\theta_{\Sigma}$. For a batch of input datapoints, those layers learn the mean and the variance of its distribution in the latent space.
 
Similarly to our work, interpolation of latent variables of VAE has been done for single phonemes \cite{hsu2017learning}. However, the previous attempt is based on very short audio chunks projected in a 128-dimensional latent space and does not provide any new strategy besides of applying VAE to audio. On the contrary, our work combines many strategies inside the network, such as dilation layers block, a significant larger compression rate at the bottleneck and, in particular, the additional classification loss at the bottleneck, besides the introduction of the MFCC-DTW as validation method.
Few examples of Variational Autoencoder fine-tuned with GANs have been studied\cite{dtu,vae_gan} for portrait images reconstruction and changing of visual attributes mapped from the latent representations.

A timbre space has been developed by means of a VAE regularized with perceptual ranking\cite{esling}. The work presents the mapping of single frames of spectral magnitude distribution computed on 2200 audio samples are mapped in a 64-dimension latent space through a 3-layer encoder with 2000 units per layer. The paths inside the latent space, where each point represents a spectral frame, are mapped by the decoder to a magnitude distribution from which it is possible to recover the phase information and the corresponding waveform. On the contrary, our work presents an end-to-end approach mapping the latent points directly to a waveform. Independently, in both works a non-synchronized decrease of the losses have been noticed. In order to overcome this issue, in \cite{esling} introduced a scaled (0.1) variational regularizer after the first 1000 epochs and null before.

\section{Autoencoder Architecture}
\begin{figure}
    \centering
    \includegraphics[width=0.45\textwidth]{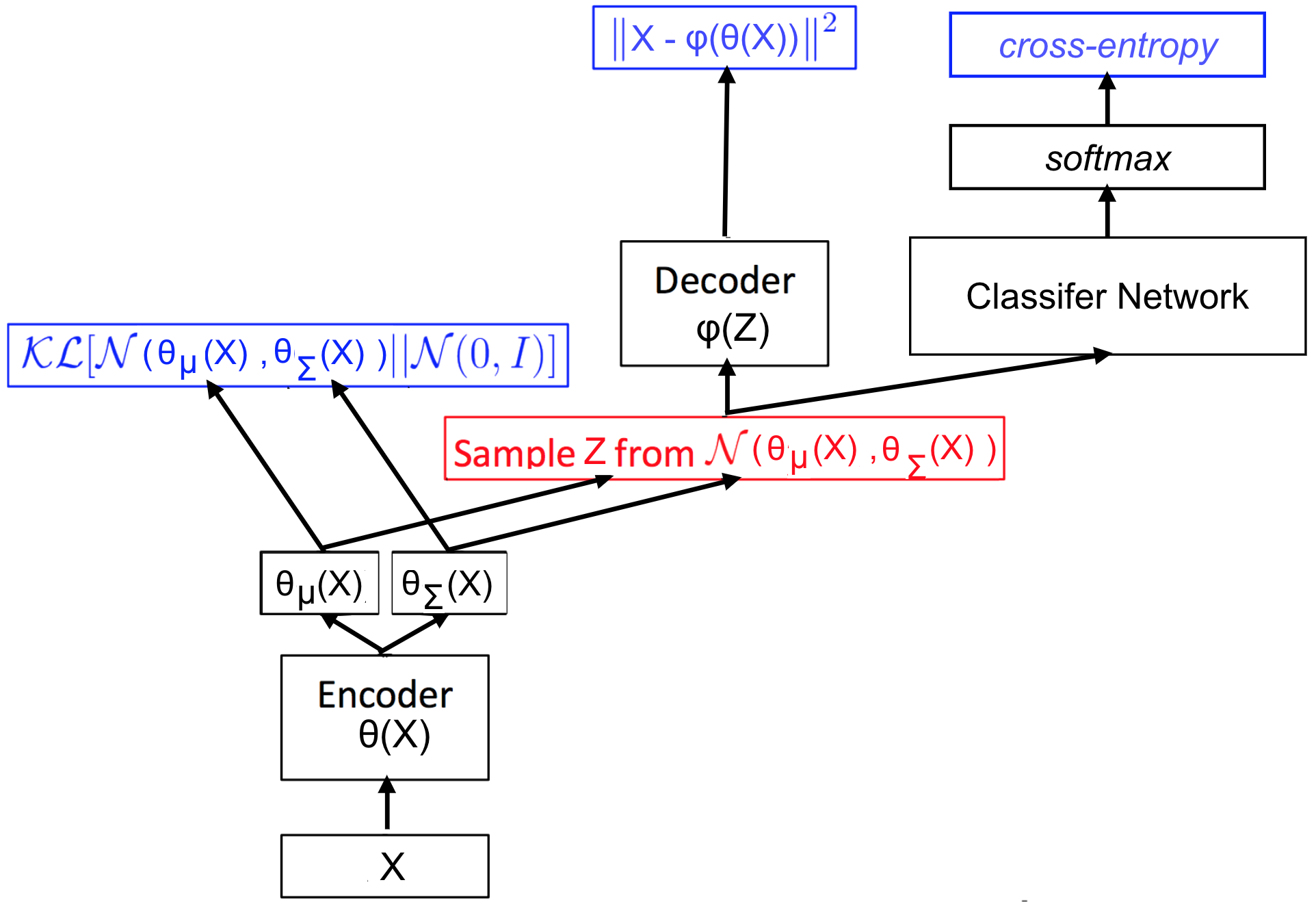}
    \caption{Structure of the VAE adding a classification loss at the bottleneck. The blue boxes show the loss functions used across the network. The encoder and decoder are structured according to table 1 and table 2. The z vector representation is sampled from the encoder applied to the input and it calculates the coordinates of the input projected into the latent space (bottleneck). (Image modified from \cite{tut_var}, p.~10.)}
    \label{fig:graph}
\end{figure}
Two autoencoders with different architectures were analysed to study the basic representation that autoencoders store in their bottleneck to describe the audio input:
\begin{itemize}
    \item convolution based autoencoder (CC-VAE),
    \item dilated convolution based autoencoder (DC-VAE).
\end{itemize}
The convolution based autoencoder consists of a stack of 12 convolution layers. The first 6 layers form the encoder. The amount of filters per layer is equal or larger than the one for the previous layer, except for the last layer. This provides a tensor with 1024 channels in the last layer before the bottleneck. The kernel size is equal to 5 for the first layer while being fixed to 4 for the rest of them. Each of the first 5 layers has a stride equal to 2. By doing so, the length of the tensor at the bottleneck is decreased by $2^{5}=32$.
The bottleneck consists of a $1\times 1$ convolution layer, collapsing the 1024 channels into a single channel. The architecture of the decoder is exactly symmetrical to the encoder.

Inspired by skip-connections, residual layers and the dilated layers used in the WaveNet, the architecture of our autoencoder is outlined in Table~\ref{table:dilation_wavenet} and~\ref{table:baseline_wavenet}. Dilated convolutions\cite{dilation} apply a filter to the input with defined gaps letting the receptive field grow exponentially by stacking them, while growing the number of parameters linearly.
The representation of the input in the bottleneck layer provides a compression by a factor of
$2^{5}=32$.
 After a first convolution layer (16 filters 32 samples long, with stride 1), a block of dilation layers
 is applied followed by 5 layers of $1 \times 2$ filters
 with a stride equal to 2. The rest of the architecture consists of a  decoder symmetrical to the encoder.
The dilation block consists of 50 residual layers \cite{res_layers} each of them composed of 2 parallel sub-layers $\text{sub\_layer}_{i=1:50}^{1,2}$ of $32 \times 2$ filters affected by a dilation factor. The dilation factor applied to the $i_{th}$ layer is $2^{mod(i-1,m)}$, where $m$ is respectively equal to 10 and to 5 for the two parallel sub-layers.
The receptive field seen by one output neuron from the last, 50th layer of the dilation block
is equal to $recf = \sum_{i=1}^{N=50}2^{mod(i-1,m)}+1= \frac{N}{m} (2^{m}-1)+1 $, $recf \mid_{m=10} = 5119$, $recf \mid_{m=5}=319$. The input of the following $i_{th}$ layer in the dilated block is given by $$\text{input}_{i}=\text{input}_{i-1}+\text{sub\_layer}_{i-1}^{1}\cdot \text{sub\_layer}_{i-1}^{2}$$
The output of the entire block is given by the average of all the 50 multiplications of sublayer outputs from each layer: $$\text{output\_block}=\frac{1}{50}\sum_{i=1}^{N=50}\text{skip\_output}_{i},$$ with
$\text{skip\_output}_{i}=\text{sub\_layer}_{i-1}^{1}\cdot \text{sub\_layer}_{i-1}^{2}.$
This procedure allows the network to extract information regarding temporal patterns across the input audio.
The bottleneck is composed of two parallel dense layers, storing a 20-dimensional vector and learning respectively the mean and the variance of the data distribution in the latent space. The variance, multiplied by a normal random vector , is summed with the mean to provide a $z$ vector in the latent space sampled from $\theta_{\mu}(x)+\theta_{\Sigma}(x)^{1/2}*\epsilon$ with $\epsilon = \mathcal{N}(1,0)$, $\theta_{\mu}$ and $\theta_{\Sigma}$ the mean and variance encoder layers, and then passed to the decoder. This procedure is called reparameterization trick, which adapts the back-propagation for the learning of the target distribution.

 \begin{table}[t]
 \centering
\begin{adjustbox}{width=0.45\textwidth}
\begin{tabular}{c}
DILATION BASED AUTOENCODER:\\ \hline \hline \\
\begin{tabular}{l|l|c|c|c}
    \textbf{\#} & \textbf{layer type:} &  \textbf{filters:} & \textbf{kernel:} & \textbf{strides} \\
    1 & conv1d & 16 & 32 & 1\\
    2& 50 x dilation\_block & 32 & 2 & 1\\
    3:7& 5 x conv1d & 1& 2&2\\
    8 & conv1d & 1& 1&1\\
    9 & dense\_a dense\_b size: 20 & -&-&-\\
    10& add &-&-&-\\
    11&conv1d\_transpose & 1& 1&1\\
    12:17&5 x conv1d\_transpose & 1& 2&2\\
    18&conv1d\_transpose & 32& 1&1\\
    19&50 x dilation\_block & 32 & 2 & 1\\
    20&conv1d\_transpose & 16 & 32 & 1\\
    21&conv1d\_transpose & 1& 1&1\\
\end{tabular}\end{tabular}\end{adjustbox}
\caption{Network architecture of our autoencoder based on dilation layers.}
\label{table:dilation_wavenet}
\end{table}

 \begin{table}[t]
 \centering
\begin{adjustbox}{width=0.42\textwidth}
\begin{tabular}{c}
CONVOLUTION BASED AUTOENCODER:\\ \hline \hline \\
\begin{tabular}{l| l|c|c|c}
    \textbf{\#} & \textbf{layer type:} &  \textbf{filters:} & \textbf{kernel:} & \textbf{strides} \\
    1&conv1d & 128 & 5 & 2\\
    2&conv1d & 128 &4& 2\\
    3&conv1d & 256& 4&2\\
    4&conv1d & 512& 4&2\\
    5&conv1d & 1024&4 &2\\
    6&conv1d & 1& 1&1\\
    7& dense\_a dense\_b size: 20 & -&-&-\\
    8& add &-&-&-\\
    9&conv1d\_transpose & 1& 1&1\\
    10&conv1d\_transpose & 1024& 4 &2\\
    11&conv1d\_transpose & 512& 4&2\\
    12&conv1d\_transpose & 256& 4&2\\
    13&conv1d\_transpose & 128& 4&2\\
    14&conv1d\_transpose & 128& 5&2\\
    15&conv1d\_transpose & 1& 1&1\\
\end{tabular}\end{tabular}\end{adjustbox}
\caption{Network architecture of our autoencoder based on convlution layers.}
\label{table:baseline_wavenet}
\end{table}

\subsection{Bottleneck Classification Loss}
Previous approaches perform well in audio reconstruction but fail to organize the classes in non-overlapping regions in the latent (bottleneck) representation, which makes morphing between two sound classes by sampling the latent space more difficult.
To enforce class separability in the bottleneck layer, we extend the autoencoder loss function by a classification loss at the bottleneck. A discriminative loss term is added via an additional classifier, composed of 3 layers consisting of 10 neurons each and applied at the bottleneck. During  training, at each iteration, the model parameters are updated using both gradients alternatingly. The model adapts its weights to reach a Nash Equilibrium between the two cost functions.
The resulting loss is given by the mean squared error between original and reconstructed signal in the time domain added to the KL-divergence between the $z$ vector from the bottleneck and a Gaussian distribution $\sim\mathcal{N}(0,1)$ plus the cross-entropy loss from the classifier at the bottleneck. The choice of the number of parameters for the classification network depends on the kind of data used. In a later application for audio of music instruments (see Section 6), the classification network has been set to have only one hidden layer. The targets of the classifier are assigned accordingly to the respective digit class of the spoken digits, while in the second experiment, the datapoints of the percussive sounds dataset are assigned to the respective clusters identified by a k-means clustering applied to the MFCC of the attack of each sound.

During training, different approaches have been tried such as awarding with bonus solo-iterations for the gradient that was slower in descending. The best solution appeared to multiply the costs by constants in order to equalize the different loss curves.

    1) the gradient calculated over the variational loss (from Eqn.~\ref{vaeobj}):  $$\nabla_{1}(\lambda_{1,1}\mathcal{L}_{Likelihood}+\lambda_{1,2}\mathcal{L}_{KL}),$$
 the gradient calculated for the classification task  on the bottleneck:
    $$\nabla_{2}(\lambda_{2}\mathcal{L}_{cross-entropy}).$$
We choose the weighting coefficients for the losses as: $\lambda_{1,1}=1.0$, $\lambda_{1,2}=0.0001$ and $\lambda_{2}=1.01$;
For $\nabla_{1}$ the learning rate is set to 0.0005, while for $\nabla_{2}$ it is  0.001.
During a first attempt of training, the model could not converge by decreasing all the three losses at once. This observation led to the decision to scale the variational regularization ($\mathcal{L}_{KL}$) by a smaller factor ($\lambda_{1,2}$) than the reconstruction loss $\mathcal{L}_{Likelihood}$. By following this procedure, the reconstruction loss of the
model decreases more quickly while the reduction of the variational loss starts to be significant after the first few tens epochs (e.g. during the training of the spoken digit model, it started to decrease between the 10th and 30th epoch). The variational loss has two main goals: the first goal is to encourage the datapoints in the latent layer to assume a multi-variate normal distribution. The second goal consists of - by means of the reparameterization trick - letting the latent space estimate the parameters of the normal distribution and associate a probabilistic representation of the original dataset to each of its latent point. This is the key-element that allows to sample the latent space and to generate valid sounds by sending a latent point between two clusters in the latent space into the decoder.

\section{Data} The dataset used consisted of is a subset of a spoken digits dataset available at \url{https://github.com/Jakobovski/free-spoken-digit-dataset}. The digits \{0,1,...,9\} have been recorded from one male speaker, repeating each digit for 50 times. The audio files consist of a mono  audio track  4096 samples long with a sample rate of 8~kHz.
The training set comprises 400 audio samples. The test set consists of 100  samples, 10 instances per digit class $i\in \{0,1,2,\ldots,9\}$.

\section{Evaluation}
The results are obtained after 117 epochs.
\begin{figure}[t]
    \centering
    \includegraphics[width=0.4\textwidth]{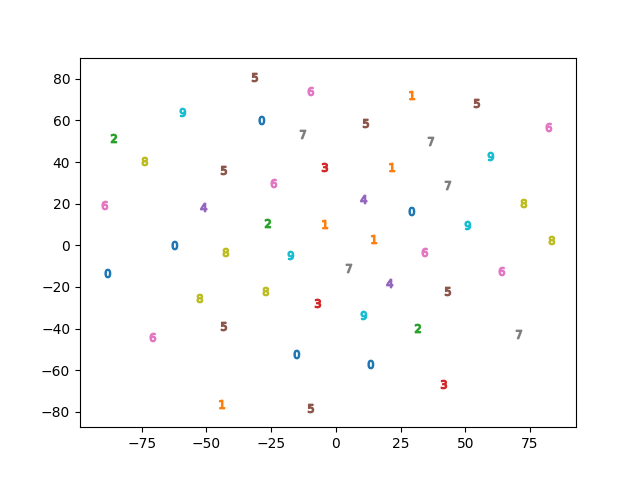}
    \centering
    \includegraphics[width=0.4\textwidth]{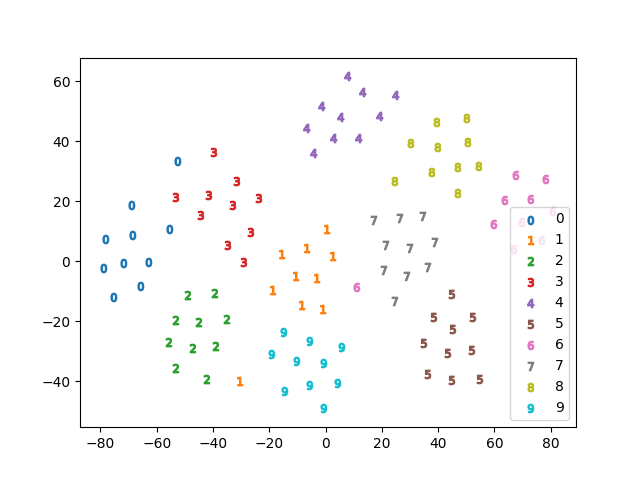}
    \caption{t-SNE representation of the test set in the 20 dimensional latent space in the dilated network without classification loss (upper image) and with classification loss (bottom image)
    VAE after introducing the loss classifier at the bottleneck.}
    \label{fig:tsne}
\end{figure}
A t-distributed Stochastic Neighbour Embedding \cite{maaten2008visualizing} (t-SNE) projection of our test set onto a 2-dimensional plane in Figure~\ref{fig:tsne} reveals the separability of the classes in the bottleneck layer.
A non-linear separability of classes in the bottleneck layer is further supported by classification accuracy of 96\% and 94\% respectively for the dilation and convolution based VAEs, when applying
1-nearest neighbour classification to the data projected into the latent space.
This indicates a good separation of the classes in the latent space.

\subsection{MFCC-DTW Deviation}
Since our goal for the latent space is to separate the audio representations into non-overlapping clusters, we are interested to have a tool in order to compare the latent datapoints belonging to the same cluster and to measure how well the decoder maps the center of a digit sound class
in the bottleneck layer to the 'center' of the audio samples of that sound class. Hereby, the center should be defined in terms of a perceptually plausible distance measure.
In order to account for non-linearly warped, yet perceptually similar sounds, the MFCC-DTW deviation uses Dynamic Time Warping (DTW)\cite{dtw}. DTW defines a measure of similarity between two signal by means of non-linear warping. We apply DTW to audio examples when representing both sounds as a sequence of Mel-frequency cepstrum coefficient (MFCC) vectors. We first calculate the DTW distance between one sound and the sound generated from the center of that sound class in the latent (bottleneck) space. Let ${\cal X}^{i}$ be the set of all sound samples of digit $i$. Be $x_{ij} \in {\cal X}^{i}$ the $j$th sample of digit class $i$.
For encoder network $\psi$, $z_{ij}=\psi(x_{ij})$ is the  projection of $x_{ij}$ into the latent space.
By encoding ${\cal X}^{i}$ via $\psi$, we yield ${\cal X}^{i}$'s projection into the latent space,
${\cal Z}^{i}=\psi({\cal X}^{i})$ where ${\cal Z}^{i}$ is the set of all the sound samples of digit $i$ encoded in the latent space.
Be $\mu({\cal Z}^{i})$ the mean of the projections of digit $i$ into the latent space. For decoder network $\theta$,
 $\theta(\mu({\cal Z}^{i}))$ is the sound generated from $\mu({\cal Z}^{i})$ by the decoder.
For a sound $x$, be $\text{mfcc}(x)$ the sequence of MFCC vectors calculated via blockwise processing of $x.$
Then we define the MFCC-DTW distance between sound sample $x_{ij}$ and the decoded latent class mean $\theta(\mu({\cal Z}^{i}))$ as
\begin{equation*}
    \text{dtw}^{\mu}_{ij} = \text{dtw}(\text{mfcc}(x_{ij}),\text{mfcc}(\theta(\mu({\cal Z}^{i}))).
\end{equation*}

Be ${\cal X}^{i}\setminus \{x_{ij}\}$  the set of samples in digit class $i$, excluding $x_{ij}$.
Then the set of MFCC-DTW distances between $x_{ij}\in {\cal X}^{i}$ and all other sounds of digit class $i$ is defined as
\begin{equation*}
    {\cal D}^{dtw}_{ij}  = \text{dtw}(\text{mfcc}(x_{ij}),\text{mfcc}({\cal X}^{i}\setminus \{x_{ij}\})).
\end{equation*}

Be $\mu({\cal D}^{dtw}_{ij})$ the mean MFCC DTW distance between sound sample $x_{ij}$  and any other sample in digit class $i$.
Be $\sigma({\cal D}^{dtw}_{ij})$ the standard deviation of ${\cal D}^{dtw}_{ij}$.
 Then we define the MFCC-DTW deviation as the difference between $\text{dtw}^{\mu}_{ij}$
and $\mu({\cal D}^{dtw}_{ij})$ normalized by the standard deviation $\sigma({\cal D}^{dtw}_{ij})$:

\begin{equation}
\text{dev}^{dtw}_{ij} = \frac{\text{dtw}^{\mu}_{ij}-\mu({\cal D}^{dtw}_{ij})}{\sigma({\cal D}^{dtw}_{ij})}.
\label{eqn:joint}
\end{equation}

For each digit class $i$, the average
$\text{dev}^{dtw}_{ij}$ is a measure of how relatively close $x_{ij}$ is to the center $\psi(\mu({\cal X}^{i}))$  of the region for that particular digit $i$. $\text{dev}^{dtw}_{ij}$ can be positive or negative. The lower $\text{dev}^{dtw}_{ij}$, the relatively closer  are $\text{dtw}^{\mu}_{ij}$ and $\mu({\cal D}^{dtw}_{ij}).$ For one digit class $i,$ the mean of $\text{dev}^{dtw}_{ij}$ across all instances $j$ of that class is a measure of how well the center of that class is preserved under the decoding operation.
In Table ~\ref{tab:alldigitsresults}, for each digit class,  mean and standard deviation of $\text{dev}^{dtw}_{ij}$ are shown across all instances of the respective digit. A negative value means that the audio generated from the center is closer, in terms of MFCC-DTW, to each of the original spoken digits than the intra-similarities among the original digits belonging to the same class.
In preserving the center per digit class, the dilation based model DC-VAE is better than the model based on regular convolution CC-VAE. Center preservation is best for digits one  and five  and worst for digit three.
Accordingly to Fig. \ref{fig:tsne}, the columns "DC-VAE (non-class)" show that the latent space without implementing a classification loss does not cluster the digits properly.

\begin{table}[t]
\centering
\begin{adjustbox}{width=0.48\textwidth}
\centering
    \begin{tabular}{l c c c c c c}
        &\multicolumn{2}{l}{DC-VAE:}&\multicolumn{2}{l}{CC-VAE:}&\multicolumn{2}{l}{DC-VAE}\\
        &&&&&\multicolumn{2}{l}{(non-class):}
        \\\hline\hline\\
    \textbf{Digit cluster:} & \textbf{Mean:} & \textbf{Std:}& \textbf{Mean:} & \textbf{Std:}& \textbf{Mean:} & \textbf{Std:}
    \\ \hline
        zero & -0.29 & 0.96 & 0.79 & 1.65& 3.39 & 2.23 \\\hline
        one  & -0.47 & 0.16 & 0.79 & 1,72  & 5.98 & 3.25 \\\hline
        two  & -0.38 & 0.51 & 0.03 & 1.13& 3.81 & 1.98  \\\hline
        three& 0.00 & 1.19 & 0.69 & 1.71&5.59 & 3.13  \\\hline
        four & -0.25 & 0.71 & 0.38 & 1.43 & 2.55 & 1.86 \\\hline
        five & -0.46 & 0.14 & 0.02 & 1.14 & 5.17 & 2.58\\\hline
        six  & -0.32 & 0.73 & -0.24& 0.93& 5.95 & 2.88 \\\hline
        seven& -0.37 & 0.27 & -0.21& 0.90 & 3.48 &  2.00 \\\hline
        eight& -0.42 & 0.47 & -0.05& 1.19 & 2.88 & 2.00\\\hline
        nine & -0.32 & 0.68 & -0.23& 0.83 & 6.41 & 4.73 \\\hline
        \textbf{average}:& -0.33 & 0.58 & 0.20 &  1.26&4.52&2.66\\
    \end{tabular}
    \end{adjustbox}
    \caption{The average MFCC-DTW deviations (Eqn.~\ref{eqn:joint}) and the standard deviations on the test set per each class are shown for the VAE with dilated convolutions (DC-VAE, columns 2 and 3), with regular convolutions (CC-VAE, columns 4 and 5) and with dilated convolutions before introducing the classification at the bottleneck (DC-VAE (non class), columns 6 and 7). The average MFCC-DTW deviations are always lower for DC-VAE compared to CC-VAE. This indicates that DC-VAE is better than CC-VAE in preserving the center of each digit class via the decoding operation.
    }
    \label{tab:alldigitsresults}
\end{table}

\section{Morphing Speech and Percussive Sounds}

The code and the generated audio samples for this project are available at \url{https://github.com/mlionello/Audio-VAE}. The repository includes the audio samples obtained for the speech morphing  between spoken digits, available at  \href{https://drive.google.com/open?id=1nSzIpJjG6bRycSS0i-70chAImWN0PVcN}{\underline{http://tiny.cc/0q92dz}}. Morphing between two digits is performed by feeding  the decoder with the latent coordinates along the path from the center of one cluster (class) of spoken digits to the center of the cluster of another spoken digit class.
The audio samples generated from the centers of each spoken digit cluster in the latent space can be found at: \href{https://drive.google.com/open?id=1Gal8SHOF0uYkFokd2dYWpTiPz8YB69L9}{\underline{http://tiny.cc/kr92dz}}.

The framework introduced in the previous section was also separately trained in the music domain, by feeding the variational autoencoder with a dataset of drums samples. The dataset contains 154 drum samples of kick, tom, hi-hat and snare sounds.
Each audio is 16384 samples long, all with sample frequency of 22~kHz. For each sound snippet, 20 MFCCs are calculated on windows of 10~ms
covering the first 70ms. The MFCCs are then averaged across the entire sound snippet.
k-means clustering  has been applied to the MFCCs in order to assign them to 5 clusters which are then identified as the 5 drum classes.
Thereby, samples with similar acoustical features are grouped in the same region of the latent space.
The dataset presents multiple challenges. Each audio file is much longer than the speech  samples used previously and each drum class contains a wide range of different waveforms.

To avoid overfitting, the classification network at the bottleneck was reduced to 1 hidden layer of size equals to the number of clusters.
The dimension of the hidden space is fixed to 30.
\begin{figure}[t]
    \centering
    \includegraphics[width=0.48\textwidth]{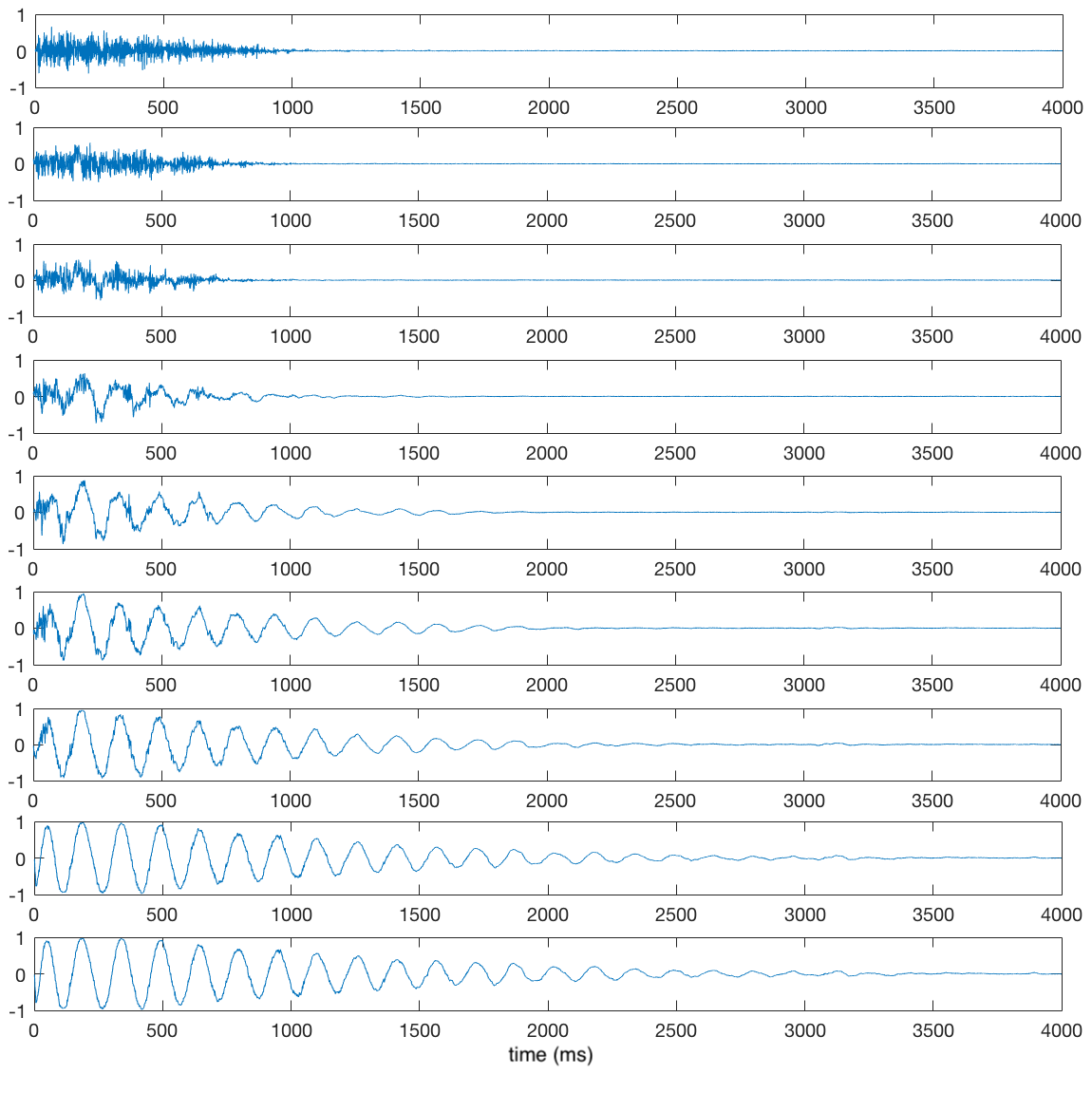}
    \caption{Morphing a hi-hat  into a tom, displayed in the time domain.}
    \label{fig:morph}
\end{figure}

The figure \ref{fig:morph} shows the morphing between the sounds of a hi-hat and a tom in the time domain. This is obtained by decoding the sampled path starting from the latent projection of one 'hi-hat' and ending at the coordinates of a 'tom'. The audio morphing showed in  figure \ref{fig:morph} can be heard at  \href{https://drive.google.com/open?id=1lByi0iaordIs31r3VnmaalnbZoaGvOTC}{\underline{http://tiny.cc/pf92dz}} and downloaded at
\href{https://github.com/mlionello/Audio-VAE/tree/master/drums/drums_output_class5_dim30/morphing}{\underline{http://tiny.cc/sh92dz}}.

\section{Conclusion}
We presented an approach for sound morphing, based on a variant of the  variational autoencoder with dilated convolutional layers and an overall loss function, augmented by the classification loss from the representation in the latent bottleneck layer. Through the integration of the latter, we turn an originally unsupervised method into a supervised one, since we now use class label information.
The use of a classification network at the bottleneck we introduced, can be further explored by implementing several discriminative losses, also working on separate sub-dimensions of the bottleneck, in order to let the latent space represent more features at the same time, including also spectral characteristics. However, several points of our work must be further investigated: (1) The small value $\lambda_{1,2}=0.0001$ deemphasizes the variational regularization to an extend that
effectively classification and reconstruction loss dominate. However it was observed that the variational loss decreased during training. (2) The number of parameters of the decoder appears to be much larger than the size of the dataset. However 1-NN and MFCC-DTW computed on the test set and reported in Section 5 show good performance in the results. (3) The dataset used for the spoken digits is limited to only one speaker. (4) No perceptual evaluation has been performed on the generated sounds. (5) The model is limited to handle only fixed length of audio data, which limits the application of such kind of models to a very restricted domain in real world applications. (6) Mean squared distance in the time domain makes the distance highly sensitive to perceptually irrelevant phase shifts between compared signals.
Finally, in future work, in order to account for  time-warped but perceptually similar audio signals,  we would like to integrate a differentiable DTW variant (such as Soft-DTW \cite{dtw_diff}) into the overall loss function.

\begin{acknowledgments}
Thanks to NVIDA for generously donating a GPU, to Tom{\'a}\v{s} Gajarsk{\'y} for helpful discussions and anonymous reviewers.
\end{acknowledgments}

\bibliography{smc2019template}

\begin{thebibliography}{10}
\providecommand{\url}[1]{#1}
\csname url@samestyle\endcsname
\providecommand{\newblock}{\relax}
\providecommand{\bibinfo}[2]{#2}
\providecommand{\BIBentrySTDinterwordspacing}{\spaceskip=0pt\relax}
\providecommand{\BIBentryALTinterwordstretchfactor}{4}
\providecommand{\BIBentryALTinterwordspacing}{\spaceskip=\fontdimen2\font plus
\BIBentryALTinterwordstretchfactor\fontdimen3\font minus
  \fontdimen4\font\relax}
\providecommand{\BIBforeignlanguage}[2]{{%
\expandafter\ifx\csname l@#1\endcsname\relax
\typeout{** WARNING: IEEEtran.bst: No hyphenation pattern has been}%
\typeout{** loaded for the language `#1'. Using the pattern for}%
\typeout{** the default language instead.}%
\else
\language=\csname l@#1\endcsname
\fi
#2}}
\providecommand{\BIBdecl}{\relax}
\BIBdecl

\bibitem{thesis}
M.~Lionello, ``A variational autoencoder approach for representation and
  transformation of sound - a deep learning approach to study the latent
  representation of sounds and to generate new audio samples,'' Master's
  thesis, Aalborg University Copenhagen, Copenhagen, Denmark, 2018.

\bibitem{Xenakis}
I.~Xenakis, \emph{Formalized Music Thought and Mathematics in
  Composition}.\hskip 1em plus 0.5em minus 0.4em\relax Indiana University
  Press, 1971.

\bibitem{chowning}
\BIBentryALTinterwordspacing
J.~M. Chowning, ``The synthesis of complex audio spectra by means of frequency
  modulation,'' \emph{J. Audio Eng. Soc}, vol.~21, no.~7, pp. 526--534, 1973.
  [Online]. Available: \url{http://www.aes.org/e-lib/browse.cfm?elib=1954}
\BIBentrySTDinterwordspacing

\bibitem{wavetable}
\BIBentryALTinterwordspacing
U.~Andresen, ``A new way in sound synthesis,'' in \emph{Audio Engineering
  Society Convention 62}, Mar 1979. [Online]. Available:
  \url{http://www.aes.org/e-lib/browse.cfm?elib=2920}
\BIBentrySTDinterwordspacing

\bibitem{x}
S.~Bilbao, \emph{Numerical Sound Synthesis: Finite Difference Schemes and
  Simulation in Musical Acoustics}.\hskip 1em plus 0.5em minus 0.4em\relax Jon
  Wiley and Sons, 2009.

\bibitem{serra}
\BIBentryALTinterwordspacing
X.~Serra and J.~Smith, ``Spectral modeling synthesis: A sound
  analysis/synthesis system based on a deterministic plus stochastic
  decomposition,'' \emph{Computer Music Journal}, vol.~14, no.~4, pp. 12--24,
  1990. [Online]. Available: \url{http://www.jstor.org/stable/3680788}
\BIBentrySTDinterwordspacing

\bibitem{purwins2019deep}
H.~Purwins, B.~Li, T.~Virtanen, J.~Schl{\"u}ter, S.-Y. Chang, and T.~Sainath,
  ``Deep learning for audio signal processing,'' \emph{IEEE Journal of Selected
  Topics in Signal Processing}, vol.~13, no.~2, pp. 206--219, 2019.

\bibitem{donahue_synthesizing_2018}
\BIBentryALTinterwordspacing
C.~Donahue, J.~McAuley, and M.~Puckette, ``Synthesizing {Audio} with
  {Generative} {Adversarial} {Networks},'' \emph{arXiv:1802.04208 [cs]}, Feb.
  2018, arXiv: 1802.04208. [Online]. Available:
  \url{http://arxiv.org/abs/1802.04208}
\BIBentrySTDinterwordspacing

\bibitem{nsynth}
\BIBentryALTinterwordspacing
J.~Engel, C.~Resnick, A.~Roberts, S.~Dieleman, D.~Eck, K.~Simonyan, and
  M.~Norouzi, ``Neural audio synthesis of musical notes with wavenet
  autoencoders,'' in \emph{34th International Conference on Machine Learning},
  2017. [Online]. Available: \url{http://arxiv.org/abs/1704.01279}
\BIBentrySTDinterwordspacing

\bibitem{wavenet}
A.~{van den Oord}, S.~{Dieleman}, H.~{Zen}, K.~{Simonyan}, O.~{Vinyals},
  A.~{Graves}, N.~{Kalchbrenner}, A.~{Senior}, and K.~{Kavukcuoglu},
  ``{WaveNet: A Generative Model for Raw Audio},'' \emph{ArXiv e-prints}, Sep.
  2016.

\bibitem{mehri_samplernn2016}
\BIBentryALTinterwordspacing
S.~Mehri, K.~Kumar, I.~Gulrajani, R.~Kumar, S.~Jain, J.~Sotelo, A.~Courville,
  and Y.~Bengio, ``{SampleRNN}: {An} unconditional end-to-end neural audio
  generation model,'' in \emph{International Conference on Learning
  Representations}, 2017. [Online]. Available:
  \url{https://openreview.net/forum?id=SkxKPDv5xl}
\BIBentrySTDinterwordspacing

\bibitem{Variational0}
\BIBentryALTinterwordspacing
D.~P. {Kingma} and M.~{Welling}, ``{Auto-Encoding Variational Bayes},'' in
  \emph{International Conference on Learning Representations}, 2017. [Online].
  Available: \url{http://adsabs.harvard.edu/abs/2013arXiv1312.6114K}
\BIBentrySTDinterwordspacing

\bibitem{hsu2017learning}
W.-N. Hsu, Y.~Zhang, and J.~Glass, ``Learning latent representations for speech
  generation and transformation,'' in \emph{Interspeech}, 2017, pp. 1273--1277.

\bibitem{gansSRNN}
\BIBentryALTinterwordspacing
V.~Kuleshov, S.~Z. Enam, and S.~Ermon, ``Audio super resolution using neural
  networks,'' \emph{CoRR}, vol. abs/1708.00853, 2017. [Online]. Available:
  \url{http://arxiv.org/abs/1708.00853}
\BIBentrySTDinterwordspacing

\bibitem{Variational1}
\BIBentryALTinterwordspacing
D.~J. Rezende, S.~Mohamed, and D.~Wierstra, ``Stochastic backpropagation and
  approximate inference in deep generative models,'' in \emph{Proceedings of
  the 31st International Conference on Machine Learning}, ser. Proceedings of
  Machine Learning Research, vol.~32, no.~2.\hskip 1em plus 0.5em minus
  0.4em\relax Bejing, China: PMLR, 22--24 Jun 2014, pp. 1278--1286. [Online].
  Available: \url{http://proceedings.mlr.press/v32/rezende14.html}
\BIBentrySTDinterwordspacing

\bibitem{pixelcnn}
\BIBentryALTinterwordspacing
A.~van~den Oord, N.~Kalchbrenner, L.~Espeholt, k.~kavukcuoglu, O.~Vinyals, and
  A.~Graves, ``Conditional image generation with pixelcnn decoders,'' in
  \emph{Advances in Neural Information Processing Systems 29}.\hskip 1em plus
  0.5em minus 0.4em\relax Curran Associates, Inc., 2016, pp. 4790--4798.
  [Online]. Available:
  \url{http://papers.nips.cc/paper/6527-conditional-image-generation-with-pixelcnn-decoders.pdf}
\BIBentrySTDinterwordspacing

\bibitem{pixelrnn}
\BIBentryALTinterwordspacing
A.~V. Oord, N.~Kalchbrenner, and K.~Kavukcuoglu, ``Pixel recurrent neural
  networks,'' in \emph{Proceedings of The 33rd International Conference on
  Machine Learning}, ser. Proceedings of Machine Learning Research,
  vol.~48.\hskip 1em plus 0.5em minus 0.4em\relax New York, New York, USA:
  PMLR, 20--22 Jun 2016, pp. 1747--1756. [Online]. Available:
  \url{http://proceedings.mlr.press/v48/oord16.html}
\BIBentrySTDinterwordspacing

\bibitem{parallelwavenet}
\BIBentryALTinterwordspacing
A.~van~den Oord, Y.~Li, I.~Babuschkin, K.~Simonyan, O.~Vinyals, K.~Kavukcuoglu,
  G.~van~den Driessche, E.~Lockhart, L.~Cobo, F.~Stimberg, N.~Casagrande,
  D.~Grewe, S.~Noury, S.~Dieleman, E.~Elsen, N.~Kalchbrenner, H.~Zen,
  A.~Graves, H.~King, T.~Walters, D.~Belov, and D.~Hassabis, ``Parallel
  {W}ave{N}et: Fast high-fidelity speech synthesis,'' in \emph{Proceedings of
  the 35th International Conference on Machine Learning}, ser. Proceedings of
  Machine Learning Research, vol.~80.\hskip 1em plus 0.5em minus 0.4em\relax
  Stockholmsmässan, Stockholm Sweden: PMLR, 10--15 Jul 2018, pp. 3918--3926.
  [Online]. Available: \url{http://proceedings.mlr.press/v80/oord18a.html}
\BIBentrySTDinterwordspacing

\bibitem{sing}
\BIBentryALTinterwordspacing
A.~Defossez, N.~Zeghidour, N.~Usunier, L.~Bottou, and F.~Bach, ``Sing:
  Symbol-to-instrument neural generator,'' in \emph{Advances in Neural
  Information Processing Systems 31}.\hskip 1em plus 0.5em minus 0.4em\relax
  Curran Associates, Inc., 2018, pp. 9055--9065. [Online]. Available:
  \url{http://papers.nips.cc/paper/8118-sing-symbol-to-instrument-neural-generator.pdf}
\BIBentrySTDinterwordspacing

\bibitem{lstm}
\BIBentryALTinterwordspacing
S.~Hochreiter and J.~Schmidhuber, ``Long short-term memory,'' \emph{Neural
  Computation}, vol.~9, no.~8, pp. 1735--1780, 1997. [Online]. Available:
  \url{https://doi.org/10.1162/neco.1997.9.8.1735}
\BIBentrySTDinterwordspacing

\bibitem{bishop_disc}
J.~Bernardo, M.~J~Bayarri, J.~O~Berger, A.~P~Dawid, D.~Heckerman, A.~F~M~Smith,
  M.~West, C.~M~Bishop, and J.~Lasserre, ``Generative or discriminative?
  getting the best of both worlds,'' \emph{BAYESIAN STATISTICS}, vol.~8, pp.
  3--24, 01 2007.

\bibitem{generative}
F.~d.~J. Thijs~Westerveld, Arjen de~Vries, ``Generative probabilistic models,''
  \emph{Multimedia Retrieval}, pp. 177--198, 2007.

\bibitem{NIPS2001_2020}
A.~Y. Ng and M.~I. Jordan, ``On discriminative vs. generative classifiers: A
  comparison of logistic regression and naive bayes,'' in \emph{Advances in
  Neural Information Processing Systems 14}, T.~G. Dietterich, S.~Becker, and
  Z.~Ghahramani, Eds.\hskip 1em plus 0.5em minus 0.4em\relax MIT Press, 2002,
  pp. 841--848.

\bibitem{fb}
N.~{Mor}, L.~{Wolf}, A.~{Polyak}, and Y.~{Taigman}, ``{A Universal Music
  Translation Network},'' \emph{ArXiv e-prints}, May 2018.

\bibitem{invgans}
A.~Creswell and A.~Anthony~Bharath, ``Inverting the generator of a generative
  adversarial network,'' \emph{IEEE Transactions on Neural Networks and
  Learning Systems}, vol.~PP, 11 2016.

\bibitem{semgans}
C.~Donahue, A.~Balsubramani, J.~McAuley, and Z.~Lipton, ``Semantically
  decomposing the latent spaces of generative adversarial networks,'' \emph{The
  sixth International Conference on Machine Learning}, 2018.

\bibitem{bigans}
\BIBentryALTinterwordspacing
D.~Bang and H.~Shim, ``High quality bidirectional generative adversarial
  networks,'' \emph{CoRR}, vol. abs/1805.10717, 2018. [Online]. Available:
  \url{http://arxiv.org/abs/1805.10717}
\BIBentrySTDinterwordspacing

\bibitem{sparse}
A.~{Makhzani} and B.~{Frey}, ``{k-Sparse Autoencoders},'' \emph{arXiv
  e-prints}, p. arXiv:1312.5663, Dec 2013.

\bibitem{tut_var}
C.~{Doersch}, ``{Tutorial on Variational Autoencoders},'' \emph{ArXiv
  e-prints}, Jun. 2016.

\bibitem{tut_var1}
\BIBentryALTinterwordspacing
R.~Ranganath, D.~Tran, J.~Altosaar, and D.~Blei, ``Operator variational
  inference,'' in \emph{Advances in Neural Information Processing Systems 29},
  D.~D. Lee, M.~Sugiyama, U.~V. Luxburg, I.~Guyon, and R.~Garnett, Eds.\hskip
  1em plus 0.5em minus 0.4em\relax Curran Associates, Inc., 2016, pp. 496--504.
  [Online]. Available:
  \url{http://papers.nips.cc/paper/6091-operator-variational-inference.pdf}
\BIBentrySTDinterwordspacing

\bibitem{vaeq}
\BIBentryALTinterwordspacing
A.~van~den Oord, O.~Vinyals, and k.~kavukcuoglu, ``Neural discrete
  representation learning,'' in \emph{Advances in Neural Information Processing
  Systems 30}, I.~Guyon, U.~V. Luxburg, S.~Bengio, H.~Wallach, R.~Fergus,
  S.~Vishwanathan, and R.~Garnett, Eds.\hskip 1em plus 0.5em minus 0.4em\relax
  Curran Associates, Inc., 2017, pp. 6306--6315. [Online]. Available:
  \url{http://papers.nips.cc/paper/7210-neural-discrete-representation-learning.pdf}
\BIBentrySTDinterwordspacing

\bibitem{dtu}
\BIBentryALTinterwordspacing
A.~B.~L. Larsen, S.~K. Sønderby, H.~Larochelle, and O.~Winther, ``Autoencoding
  beyond pixels using a learned similarity metric,'' in \emph{Proceedings of
  The 33rd International Conference on Machine Learning}, ser. Proceedings of
  Machine Learning Research, M.~F. Balcan and K.~Q. Weinberger, Eds.,
  vol.~48.\hskip 1em plus 0.5em minus 0.4em\relax New York, New York, USA:
  PMLR, 20--22 Jun 2016, pp. 1558--1566. [Online]. Available:
  \url{http://proceedings.mlr.press/v48/larsen16.html}
\BIBentrySTDinterwordspacing

\bibitem{vae_gan}
\BIBentryALTinterwordspacing
L.~Mescheder, S.~Nowozin, and A.~Geiger, ``Adversarial variational bayes:
  Unifying variational autoencoders and generative adversarial networks,'' in
  \emph{Proceedings of the 34th International Conference on Machine Learning},
  ser. Proceedings of Machine Learning Research, D.~Precup and Y.~W. Teh, Eds.,
  vol.~70.\hskip 1em plus 0.5em minus 0.4em\relax International Convention
  Centre, Sydney, Australia: PMLR, 06--11 Aug 2017, pp. 2391--2400. [Online].
  Available: \url{http://proceedings.mlr.press/v70/mescheder17a.html}
\BIBentrySTDinterwordspacing

\bibitem{esling}
P.~Esling, A.~Chemla{-}Romeu{-}Santos, and A.~Bitton, ``Generative timbre
  spaces with variational audio synthesis,'' in \emph{International Conference
  on Digital Audio Effects}, 2018.

\bibitem{dilation}
F.~Yu and V.~Koltun, ``Multi-scale context aggregation by dilated
  convolutions,'' in \emph{International Conference on Learning
  Representations}, 05 2016.

\bibitem{res_layers}
\BIBentryALTinterwordspacing
K.~He, X.~Zhang, S.~Ren, and J.~Sun, ``Deep residual learning for image
  recognition,'' \emph{IEEE Conference on Computer Vision and Pattern
  Recognition}, 2016. [Online]. Available:
  \url{http://arxiv.org/abs/1512.03385}
\BIBentrySTDinterwordspacing

\bibitem{maaten2008visualizing}
L.~v.~d. Maaten and G.~Hinton, ``Visualizing data using t-sne,'' \emph{Journal
  of machine learning research}, vol.~9, no. Nov, pp. 2579--2605, 2008.

\bibitem{dtw}
H.~{Sakoe} and S.~{Chiba}, ``Dynamic programming algorithm optimization for
  spoken word recognition,'' \emph{IEEE Transactions on Acoustics, Speech, and
  Signal Processing}, vol.~26, no.~1, pp. 43--49, February 1978.

\bibitem{dtw_diff}
\BIBentryALTinterwordspacing
M.~Cuturi and M.~Blondel, ``{Soft-DTW: a Differentiable Loss Function for
  Time-Series},'' Center for Research in Economics and Statistics, Working
  Papers 2017-81, Mar. 2017. [Online]. Available:
  \url{https://ideas.repec.org/p/crs/wpaper/2017-81.html}
\BIBentrySTDinterwordspacing

\end{thebibliography}

\end{document}